\documentclass[10pt,twocolumn,letterpaper]{article}

\usepackage{iccv}
\usepackage{times}
\usepackage{epsfig}
\usepackage{graphicx}
\usepackage{amsmath}
\usepackage{amssymb}
\usepackage{bm}
\usepackage{booktabs}

\usepackage[dvipsnames]{xcolor}

\definecolor{citecolor}{RGB}{0, 113, 188}
\usepackage[pagebackref,breaklinks,colorlinks, citecolor=ForestGreen]{hyperref}
\usepackage[capitalize,noabbrev]{cleveref}

\iccvfinalcopy 

\def\iccvPaperID{****}

\ificcvfinal\fi

\begin{document}

\title{Implicit Temporal Modeling with Learnable Alignment for Video Recognition}

\author{Shuyuan Tu$^{1,2}$ \ \ \ \ 
Qi Dai$^3$ \ \ \ \ 
Zuxuan Wu$^{1,2}$ \thanks{Corresponding author} \ \ \ \
Zhi-Qi Cheng$^{4}$ \ \ \ \ 
Han Hu$^3$ \ \ \ \ 
Yu-Gang Jiang$^{1,2}$ \vspace{0.1in}\\
{$^1$Shanghai Key Lab of Intell. Info. Processing, School of CS, Fudan University} \\
{$^2$Shanghai Collaborative Innovation Center of Intelligent Visual Computing} \\
{$^3$Microsoft Research Asia}  \quad 
{$^4$Carnegie Mellon University} 
}

\maketitle
\ificcvfinal\thispagestyle{empty}\fi

\begin{abstract}
Contrastive language-image pretraining (CLIP) has demonstrated remarkable success in various image tasks. 
However, how to extend CLIP with effective temporal modeling is still an open and crucial problem. 
Existing factorized or joint spatial-temporal modeling trades off between the efficiency and performance. 
While modeling temporal information within straight through tube is widely adopted in literature, we find that simple frame alignment already provides enough essence without temporal attention. 
To this end, in this paper, we proposed a novel Implicit Learnable Alignment (ILA) method, which minimizes the temporal modeling effort while achieving incredibly high performance. 
Specifically, for a frame pair, an interactive point is predicted in each frame, serving as a mutual information rich region. 
By enhancing the features around the interactive point, two frames are implicitly aligned. 
The aligned features are then pooled into a single token, which is leveraged in the subsequent spatial self-attention. 
Our method allows eliminating the costly or insufficient temporal self-attention in video. 
Extensive experiments on benchmarks demonstrate the superiority and generality of our module. 
Particularly, the proposed ILA achieves a top-1 accuracy of 88.7\% on Kinetics-400 with much fewer FLOPs compared with Swin-L and ViViT-H.
Code is released at \url{https://github.com/Francis-Rings/ILA}.
\end{abstract}

\section{Introduction}

\noindent Video recognition is rated as one of the most fundamental components of video understanding. 
Numerous downstream tasks heavily rely on the basic recognition model, \emph{e.g.}, action localization \cite{cheng2017video2shop,chao2018rethinking,shi2021temporal,shi2020weakly,shou2016temporal}, detection \cite{cheng2017video,cheng2016video,hou2017tube,li2018recurrent,deng2020single}, and video object tracking \cite{he2023damo,wang2019fast,wang2022omnivl}.
Due to the great potential of video technologies, it has been an active research direction over the past few years.
Various approaches have been proposed, including convolution-based methods \cite{b1,b2,b4,b5,b9,b11,b12,wu2020dynamic} and transformer-based methods \cite{b17,b13,cheng2022gsrformer,b16,b18,b15,tian2023resformer,xing2023svformer}.
Recently, Contrastive Language-Image Pretraining (CLIP) \cite{b21} has demonstrated strong performance in video domain.
Studies \cite{b22,b23,b24,b25,pan2022st,yang2023aim,weng2023transforming} attempt to transfer the powerful CLIP model to video tasks, which promote the recognition performance to a brand-new level, showing its general representation ability.

\begin{figure}[t]
\begin{center}
\includegraphics[width=0.95\linewidth]{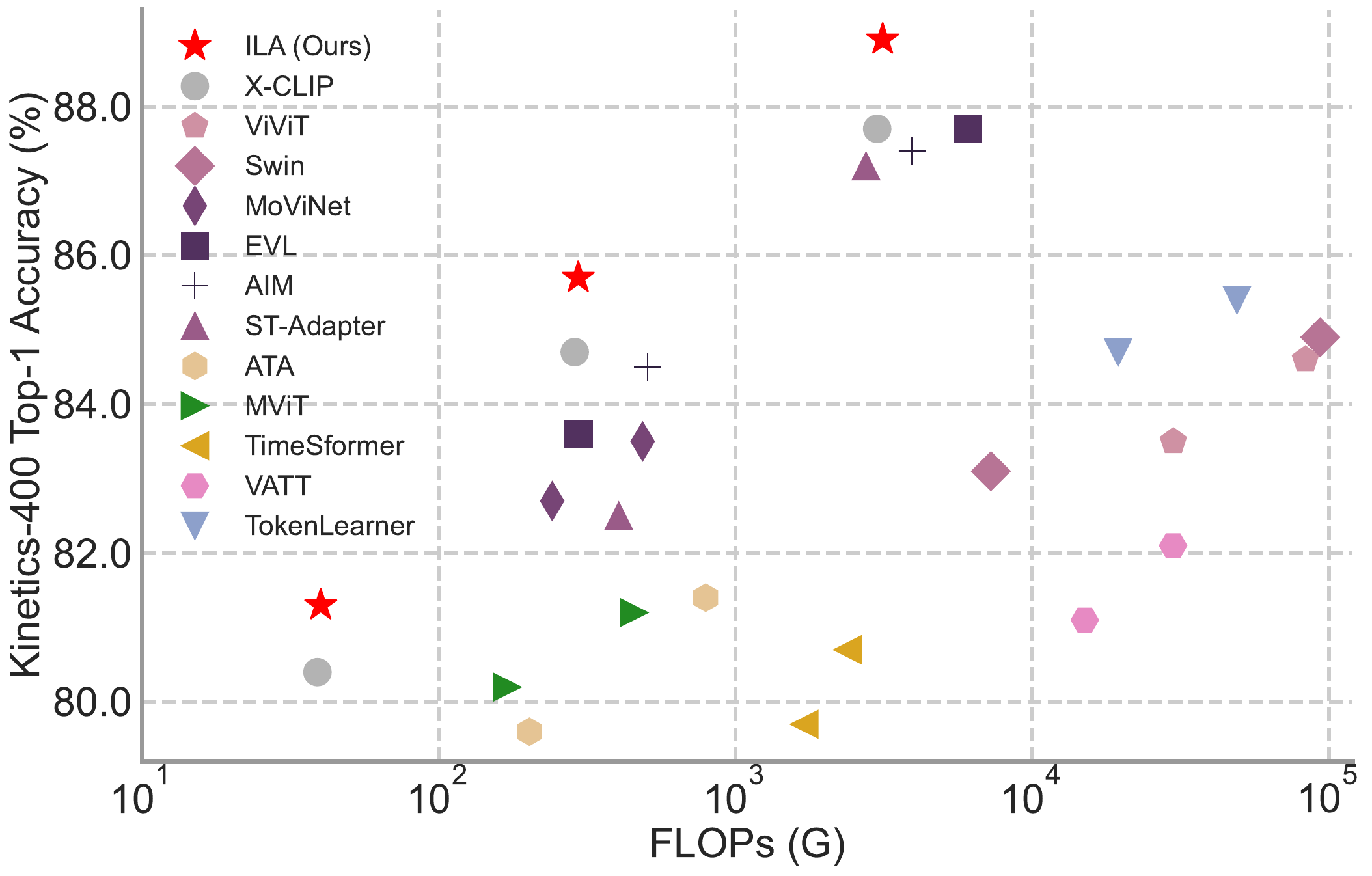}
\end{center}
   \caption{Top-1 accuracy comparison with state-of-the-art methods on Kinetics-400 \cite{b46} under different FLOPs. ILA achieves competitive results. Best viewed in color.}
\label{fig:acc_flops}
\vspace{-0.2cm}
\end{figure}

Generally, existing methods devise various temporal modeling schemes to explore the potential of CLIP, including the factorized~\cite{yang2023aim} or frame-level~\cite{b24,b23} temporal attention, and temporal cross attention~\cite{b25}.
All these tailored methods aim at designing lightweight temporal modules to reuse the CLIP model.
Though considerable improvements are achieved, such temporal modeling approaches still depend on the complex self-attention, which we argue is not necessary in CLIP-based framework.

In this paper, we rethink the role of temporal modeling in general CLIP-based video recognition framework. Unlike existing approaches rely on temporal attention, we hypothesize that important motion and action clues can be derived when performing alignment of pairwise frames.
As a result, the costly~\cite{b15,b13} or insufficient~\cite{b24,b23,b25} temporal attentions can be avoided, without harming the performance.
While explicit patch alignment is time consuming with low efficiency, we prioritize only an implicit and coarse alignment, aiming at involving the vital temporal signals.

\begin{figure}[t]
\begin{center}
\includegraphics[width=0.95\linewidth]{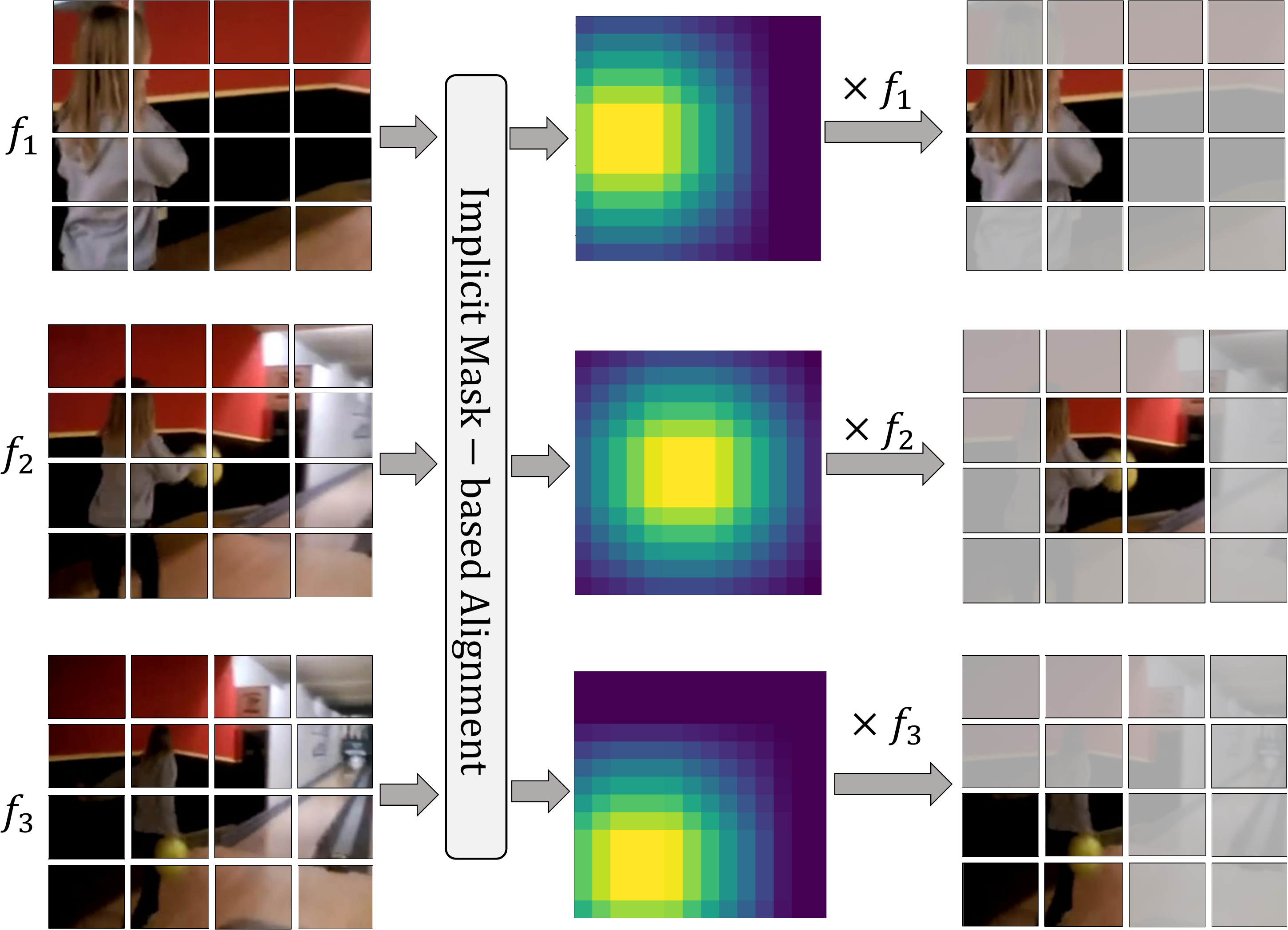}
\end{center}
   \caption{The proposed ILA employs an implicit and coarse mask to align the features, which focus on the active interaction region. We hypothesize the important motion and action clues can be derived from aligned features.}
\label{fig:comparisons}
\vspace{-0.2cm}
\end{figure}

In light of this, we present a novel Implicit Learnable Alignment (ILA) method for efficient video recognition. More specifically, ILA employs learnable masks to align features of two adjacent frames. The alignment is achieved with the help of an interactive point that is predicted using a convolution module conditioned on a frame pair. Based on the point, a corresponding region is generated indicating close interactions of adjacent frames. 
The mask is defined as the map of weights implying which region contains vital information.
We then assign higher weights around the interactive point in the mask, while assigning lower weights to other positions, suppressing irrelevant signals among them.
By leveraging the generated mask to weight the frame representations, coarsely aligned features are obtained, as shown in Figure \ref{fig:comparisons}.
Note all above operations are performed in parallel among frame pairs to boost the speed. To efficiently and fully exploit the alignment, the aligned features are pooled into a single mutual information token.
The token is subsequently concatenated with other frame tokens to perform the spatial self-attention, which implicitly models the temporal relations between frames. Our method is plugged into each spatial block of vision transformer and forms the Implicit Spatio-Temporal attention (IST) block, which allows temporal modeling without the use of the traditional temporal self-attention.

Our contributions can be summarized as follows: (1) We propose  Implicit Learnable Alignment (ILA) for video recognition. Our implicit temporal modeling can be seamlessly plugged into existing vision transformer models. It utilizes the coarse alignment as the key temporal signals, which enables superior temporal modeling at a low computational cost.
(2) We show that such a simple frame alignment already encodes the essence of temporal relations, which allow eliminating the insufficient temporal self-attention.
(3) Extensive qualitative and quantitative experiments demonstrate the effectiveness and efficiency of ILA. We achieve 88.7\% on Kinetics-400 with low computation overhead. Our method builds a promising bridge for CLIP from image processing to video recognition.

\section{Related Work}

\noindent \textbf{Visual-language representation learning} has demonstrated remarkable success in various tasks \cite{b21,jia2021scaling,b19}. By leveraging contrastive learning between language and image, a joint representation space is learned.
Particularly, CLIP~\cite{b21} has shown its strong power in open domain problems, and dozens of approaches are developed, including few-shot learning \cite{b29,b30}, point cloud understanding \cite{b31,b32}, video understanding \cite{b33,b22,b23}, \emph{etc.}

Recently, several studies extend the existing CLIP model to the video domain.
X-CLIP \cite{b24} devises the frame-level temporal attention to avoid high computation.
EVL \cite{b25} employs temporal convolution and cross-attention on top of the CLIP features.
ST-Adapter \cite{pan2022st} inserts the spatiotemporal adapter into each block, which consists of several 3D convolution layers.
AIM \cite{yang2023aim} reuses the CLIP self-attention as the temporal ones via an additional adapter module.
Nevertheless, the above methods explore lightweight adaptations of CLIP using insufficient temporal attention, \emph{e.g.}, frame-level or local temporal attention.
In our work, we attempt to perform temporal modeling with signals emerged from a simple alignment process, which involves the comprehensive temporal clues yet remains simplicity.

\vspace{0.05in}
\noindent \textbf{Video recognition} is the key task in video understanding. In the convolution era, two-stream networks \cite{b1,b2,b3} and spatiotemporal CNNs \cite{b4,b7,b8,b9} are proposed. The former treats spatial representations and optical flow images as two independent modules, and the latter employs (separable) 3D convolution to extract spatiotemporal features.
Recently, inspired by vision transformers~\cite{dosovitskiy2020image,touvron2021training,zhanghivit,tian2023resformer,han2021connection}, video transformers \cite{b13,b15,b16,b17,b37,akbari2021vatt} have shown promising results compared to CNN methods, due to their much larger receptive fields.
TimeSformer \cite{b13} adopts factored space time attention as a trade-off between speed and accuracy.
ViViT \cite{b17} investigates four types of temporal attention, and selected the global spatiotemporal attention as the default.
Video Swin \cite{b15} uses local spatiotemporal attention to model the temporal information.
However, these methods are either computationally intensive or insufficient in modeling the temporal interactions, resulting in high model cost or unsatisfactory performance.
In contrast, our method explores how to model the complex temporal information with minimal effort, demonstrating the redundancy in existing temporal attention models.

\vspace{0.05in}
\noindent \textbf{Temporal correspondences} reflect the motions in video and can be used in several video understanding tasks~~\cite{b14,b35,b36,b37,b38,b39}.
For example, in video super resolution, alignment-based methods~\cite{b40,b41,b42,b43,b44,b45} have been proposed to keep frames coherent.
PSRT-recurrent~\cite{b40} points out that patch level alignment can reduce memory cost when computing optical flow.
While in video recognition, the recent ATA~\cite{b28} adopts Hungarian matching to align the patches between frames, and then performs temporal attention within aligned patches, followed by the de-alignment.
However, the model is significantly encumbered with the slow serial alignment, followed by computationally expensive operations to calculate temporal attention. 
In contrast, our approach employs learnable masks to align frames in parallel with an aim to involve important motion and action clues, thus benefiting the video understanding. Therefore, the alignment in our method is implicit and coarse.

\section{Method}

In this section, we elaborate our proposed architecture in detail. First, we introduce the overview of our ILA in Section \ref{sec:overview}. Second, we depict the concrete implicit mask-based alignment in Section \ref{sec:align}. Finally, we describe the loss functions of our dedicated framework.

\subsection{Architecture Overview}
\label{sec:overview}

The proposed ILA model consists of several Implicit Spatio-Temporal attention (IST) blocks.
The model is built upon a pretrained image vision transformer (ViT) \cite{dosovitskiy2020image}.
While previous methods \cite{b17,b13} mostly rely on the ImageNet initialized models, recent approaches \cite{b24,b25,pan2022st,yang2023aim} have revealed the powerful representation ability of large-scale visual-language pretrained models \cite{b21,b19}.
Our method follows the predecessors and is initialized from the CLIP model \cite{b21}.
Given an input video clip $\mathbf{x}=[\mathbf{x}_1,...,\mathbf{x}_t,...,\mathbf{x}_T], \mathbf{x}_t\in \mathbb{R}^{H\times W \times 3}$, we decompose each frame into $\frac{H}{P}\times \frac{W}{P}$ non-overlapping patches$\{\mathbf{x}_{t,i}\}^{hw}_{i=1}$, where $T, H, W$ are the number of frames, height and width, $h=\frac{H}{P}, w=\frac{W}{P}$. $P$ is the patch size.
The patches are linearly mapped to embedding vectors $\mathbf{z}^{(0)}_{t}=[\mathbf{z}^{(0)}_{t,1},...,\mathbf{z}^{(0)}_{t,i},...,\mathbf{z}^{(0)}_{t,hw}],~\mathbf{z}^{(0)}_{t,i}\in \mathbb{R}^d$:
\begin{equation} 
\label{eq:embed}
\mathbf{z}^{(0)}_{t,i}=\mathbf{E}\mathbf{x}_{t,i}+\mathbf{e}^{pos}_{t,i},
\end{equation}
where $\mathbf{E}\in \mathbb{R}^{d \times 3P^2}$ is the projection matrix, and $\mathbf{e}^{pos}_{t,i}$ is the spatial positional embedding. We also add a classification token $\mathbf{z}^{(0)}_{t,cls}$ for each frame.

\begin{figure}[t]
\begin{center}
\includegraphics[width=0.95\linewidth]{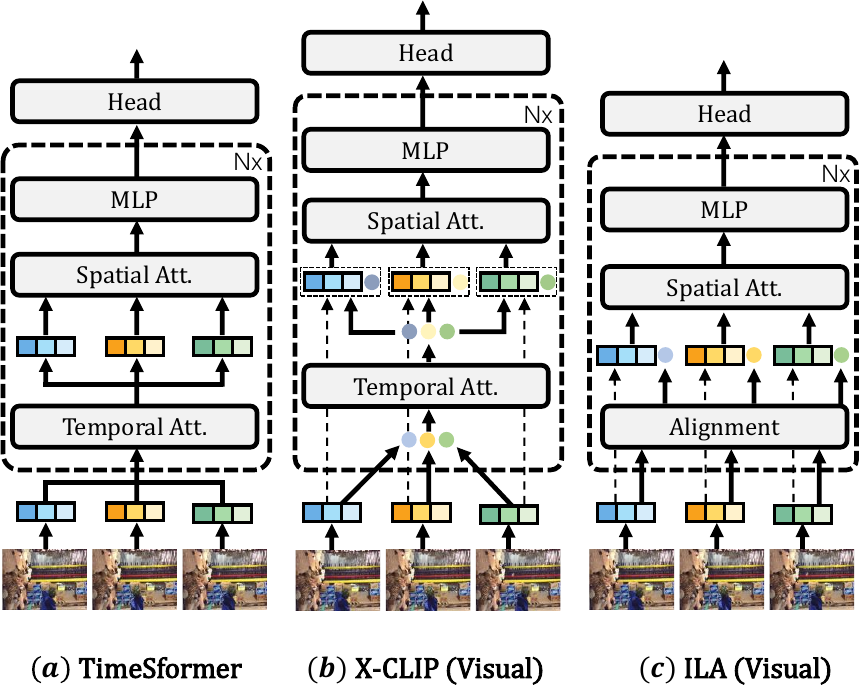}
\end{center}
\vspace{-0.2cm}
   \caption{The structures of three different models. (a) The divided spatiotemporal attention in TimeSformer~\cite{b13}. (b) The frame-level temporal attention in X-CLIP~\cite{b24}. (c) The alignment-based temporal modeling in our ILA. 
   }
\label{fig:overview}
\vspace{-0.2cm}
\end{figure}

The structure of the IST block is illustrated in Figure \ref{fig:overview}. At each IST block $\ell$, we align the semantic features of each consecutive frame pair ($\mathbf{z}^{(\ell-1)}_{t},~\mathbf{z}^{(\ell-1)}_{t-1}$) via finding an interactive position (as will be introduced in \cref{sec:align}) per frame, which serves as a mutual information (MI) rich region.
By simply weighting the feature map with higher weights surrounding the interactive position, the aligned features $\mathbf{{a}}^{(\ell)}_{t},~\mathbf{{a}}^{(\ell)}_{t-1}$ are obtained:
\begin{equation}\small
\label{eq:align}
\begin{aligned}
     \mathbf{{a}}^{(\ell)}_{t},~\mathbf{{a}}^{(\ell)}_{t-1}=\mathtt{Align}(\mathbf{z}^{(\ell-1)}_{t}, \mathbf{z}^{(\ell-1)}_{t-1}). 
\end{aligned}
\end{equation}
The aligned features are subsequently mean-pooled into a single mutual information token $\mathbf{\hat{z}}^{(\ell)}_{t,mut}$, which is further concatenated with corresponding frame to perform the spatial Multi-head Self Attention (MSA):
\begin{subequations}\small
\begin{align}
\mathbf{\hat{z}}^{(\ell)}_{t,mut}&=\mathtt{Avg}(\mathbf{{a}}^{(\ell)}_{t}),\label{eq:pool}\\
[\mathbf{\tilde{z}}^{(\ell)}_{t},~\mathbf{\tilde{z}}^{(\ell)}_{t,mut}] = \mathtt{MSA}(\mathtt{LN}([&\mathbf{z}^{(\ell-1)}_{t},\mathbf{\hat{z}}^{(\ell)}_{t,mut}]))+[\mathbf{z}^{(\ell-1)}_{t},\mathbf{\hat{z}}^{(\ell)}_{t,mut}], \label{eq:msa}
\end{align}
\end{subequations}
where $\mathtt{LN}(\cdot)$ indicates layer normalization \cite{ba2016layer}.
$\mathbf{\tilde{z}}^{(\ell)}_{t,mut}$ is then dropped before feeding to the MLP, and the output of block $\ell$ is formulated as:
\begin{equation}\small
\label{eq:mlp}
\begin{aligned}
     \mathbf{z}^{(\ell)}_{t} = \mathtt{MLP}(\mathtt{LN}(\mathbf{\tilde{z}}^{(\ell)}_{t})) + \mathbf{\tilde{z}}^{(\ell)}_{t}.
\end{aligned}
\end{equation}

\begin{figure*}[t!]
\begin{center}
\includegraphics[width=0.95\linewidth]{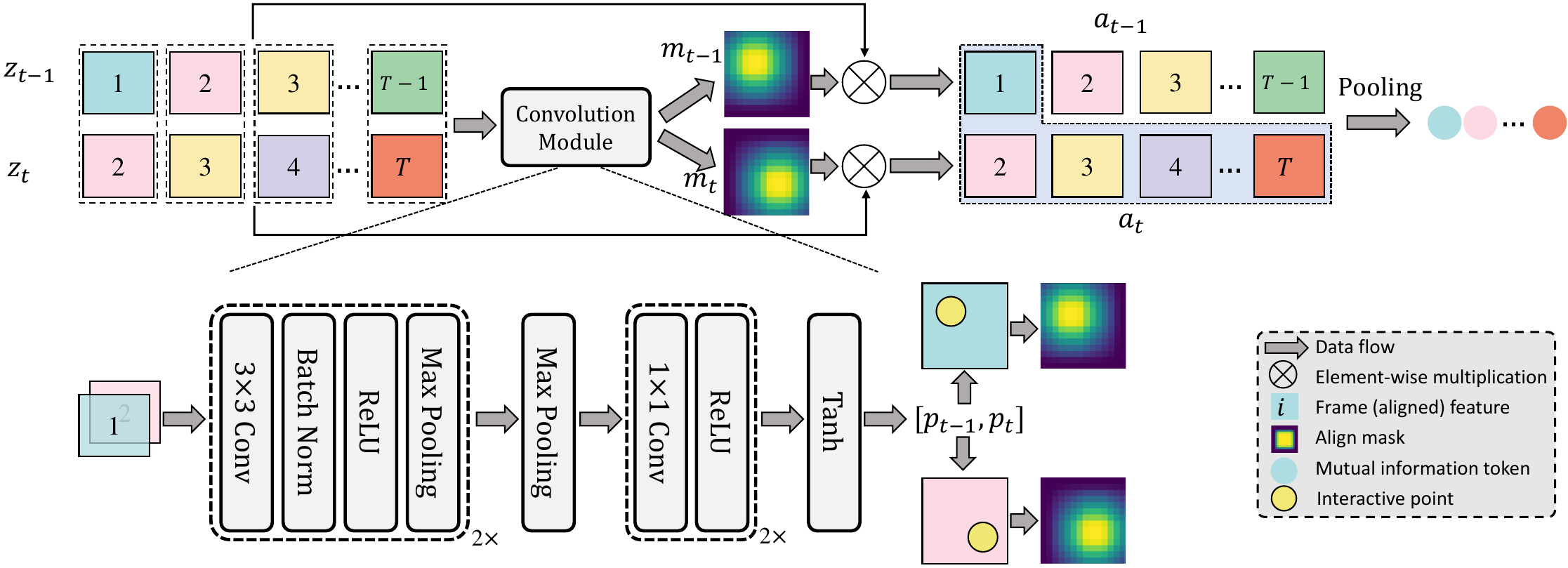}
\end{center}
\vspace{-0.3cm}
   \caption{Details of the proposed alignment method. For each adjacent frame pair, a convolution module is leveraged to predict one interactive point per frame, which refers to region with close interactions between frames.
   A mask is generated by assigning higher weights around the interactive point, while assigning lower weights to other positions.
   The mask is then adopted to weight the frame features, obtaining aligned features.
   Finally, the aligned features are pooled into a single mutual information token.
   Best viewed in color.
   }
\label{fig:ILA}
\vspace{-0.2cm}
\end{figure*}

Unlike common supervised frameworks that use one-hot labels as the target, to fully leverage the pretrained visual-language model, we follow \cite{b24} to optimize the similarity loss supervised by textual information of categories.
Formally, the text representation $\mathbf{c}$ is computed by inputting the category name to the text encoder $f_{t}(\cdot)$.
Then a video-specific prompt is obtained by querying $\mathbf{c}$ among video representation $\{\mathbf{z}^{(L)}_{t}\}^{T}_{t=1}$ ($L$ is the number of IST blocks), which is further used to enhance $\mathbf{c}$.
Finally, the model maximizes the cosine similarity between the video and text representations if they are matched, otherwise minimizes it.

\subsection{Implicit Mask-based Alignment}
\label{sec:align}
The IST block employs an implicit mask-based alignment component to align the semantic features between two frames.
A previous study \cite{b28} had explored patch-level alignment through Hungarian matching \cite{bertsekas1981new}, which however suffered from limited performance and low efficiency.
On one hand, the explicit patch alignment focuses on patch coherence across frames, which can eliminate possible beneficial temporal interactions.
On the other hand, such alignment must be operated frame by frame with cubic time complexity, incurring significant computational overhead.
In contrast, our implicit alignment attempts to enhance favorable mutual information and in turn suppress irrelevant information with learned masks.
As such, the key temporal clues are preserved while allowing flexible and efficient computation.

Figure \ref{fig:ILA} illustrates the details of our alignment method, which is concretely described as follows.
In the $\ell$-th block, we duplicate each input clip $\{\mathbf{z}^{(\ell-1)}_{t}\}^{T}_{t=1}$ to form an adjacent input pair $\{(\mathbf{z}^{(\ell-1)}_{t},\mathbf{z}^{(\ell-1)}_{t-1})\}^{T}_{t=2}$.
Each pair of representations are then concatenated along the channel dimension, which are further fed into a dedicated lightweight convolution module for predicting two interactive points:
\begin{equation}\small
\label{eq:conv}
\begin{aligned}
     {\bm{p}}^{(\ell)}_{t},~{\bm{p}}^{(\ell)}_{t-1}=\mathtt{Conv}(\mathtt{Concat}(\mathbf{z}^{(\ell-1)}_{t}, \mathbf{z}^{(\ell-1)}_{t-1})),
\end{aligned}
\end{equation}
where the convolution module $\mathtt{Conv}(\cdot)$ consists of a sequence of convolution, normalization and pooling layers.
The interactive points ${\bm{p}}^{(\ell)}_{t},~\bm{{p}}^{(\ell)}_{t-1}\in \mathbb{R}^{2}$ represent the most semantically similar positions in two frames, indicating the region with favorable mutual information.
We assume the closer the position is to the interactive point, the more temporal information it involves.
On the contrary, a position that is far away from the interactive point can contain redundant and irrelevant information, which should be suppressed.
To this end, two align masks $\mathbf{m}^{(\ell)}_{t},\mathbf{m}^{(\ell)}_{t-1}\in \mathbb{R}^{h\times w}$ are generated by endowing positions closer to the interactive points with higher weights. Formally, for a spatial position $\bm{u}$ in $\mathbf{m}^{(\ell)}_{t}$, its weight $w_{\bm{u}}$ is computed by:
\begin{equation}\small
\label{eq:mask1}
\begin{aligned}
     s = \mathtt{dist}(\bm{u},{\bm{p}}^{(\ell)}_{t}), ~~~~~~~~~~~~~~~~~~~~~~~~\\
     w_{\bm{u}}=\left\{\begin{array}{l}
     \eta,~~~~~~~~~~~~~~~~~~~~~~~~~~~~~~~~~~~~~\text{if}~~s \leq \delta, \\
     \mathtt{max} \left(0,\eta-\beta\left(s-\delta\right)\right),~~\text{if}~~s>\delta,
     \end{array}\right.
\end{aligned}
\end{equation}
where $\mathtt{dist}(\cdot)$ is the distance function, and $\eta, \delta, \beta$ are the parameters.
The weights of $\mathbf{m}^{(\ell)}_{t-1}$ are obtained by similar calculation with $\bm{{p}}^{(\ell)}_{t-1}$.
Note that all the coordinates of positions are scaled to the range $[-1,1]$ to facilitate the mask calculation.
The aligned feature representations $\mathbf{{a}}^{(\ell)}_{t},\mathbf{{a}}^{(\ell)}_{t-1}$ are produced by weighting the frame features with the align masks:
\begin{subequations}\small
\begin{align}
\mathbf{{a}}^{(\ell)}_{t}&=\mathbf{m}^{(\ell)}_{t}\mathbf{z}^{(\ell-1)}_{t},\label{eq:mul1}\\
\mathbf{{a}}^{(\ell)}_{t-1}&=\mathbf{m}^{(\ell)}_{t-1}\mathbf{z}^{(\ell-1)}_{t-1}. \label{eq:mul2}
\end{align}
\end{subequations}

We hypothesize that the aligned feature can implicitly preserve the mutual information and already encodes essential temporal information, which could be leveraged to model the temporal relations across frames.
Nevertheless, directly replacing $\mathbf{z}^{(\ell-1)}_{t}$ with the aligned feature $\mathbf{{a}}^{(\ell)}_{t}$ would prejudice the performance, since $\mathbf{{a}}^{(\ell)}_{t}$ focuses more on the interaction region while ignoring the spatial correlations.
Instead, we consider $\mathbf{{a}}^{(\ell)}_{t}$ as a specific temporal signal.
Thus, we averagely pool the feature into a single mutual information token $\mathbf{\hat{z}}^{(\ell)}_{t,mut}$ (Eq. \eqref{eq:pool}), which is further utilized in spatial multi-head self attention (Eq. \eqref{eq:msa}).
Note that since we duplicate the input clip to form frame pairs, there are two aligned features for frame $\mathbf{{z}}^{(\ell-1)}_{t}$, $2\leq t \leq T-1$.
For example, $\mathbf{{a}}^{(\ell)}_{t}$ can be computed from both pairs $(\mathbf{z}^{(\ell-1)}_{t}, \mathbf{z}^{(\ell-1)}_{t-1})$ and $(\mathbf{z}^{(\ell-1)}_{t+1}, \mathbf{z}^{(\ell-1)}_{t})$.
In our implementation, only $\mathbf{{a}}^{(\ell)}_{t}$ computed from $(\mathbf{z}^{(\ell-1)}_{t}, \mathbf{z}^{(\ell-1)}_{t-1})$ is exploited for pooling to the mutual information token.

\begin{table*}[t!]\small
\caption{Comparison with the state-of-the-arts on Kinetics-400. The FLOPs per view of each method is reported. We categorize methods by different pretraining data.}
 \vspace{-0.1in}
\begin{center}
\renewcommand\arraystretch{1.1}
\scalebox{0.9}{
\begin{tabular}{lcccccc}
\toprule
\textbf{Model}                                                  & \textbf{Pretrain}  & \textbf{Frames} & \textbf{Top-1}         & \textbf{Top-5}         & \textbf{Views} & \textbf{FLOPs~(G)} \\ \midrule
\textit{\textbf{Random initialization}}                &           &        &               &               &       &          \\
MViTv1-B~\cite{b16}                                               & -         & 64     & 81.2          & 95.1          & 3×3   & 455      \\ \midrule
\textit{\textbf{ImageNet pretraining}}                 &           &        &               &               &       &          \\
Uniformer-B~\cite{b18}                                            & IN-1K     & 32     & 83.0          & 95.4          & 4×3   & 259      \\
TimeSformer-L~\cite{b13}                                          & IN-21K    & 96     & 80.7          & 94.7          & 1×3   & 2380     \\
ATA~\cite{b28}                                                    & IN-21K    & 32     & 81.9          & 95.5          & 4×3   & 793      \\
Mformer-HR~\cite{b14}                                             & IN-21K    & 16     & 81.1          & 95.2          & 10×3  & 959      \\
Swin-L~(@384px)~\cite{b15}                                         & IN-21K    & 32     & 84.9          & 96.7          & 10×5  & 2107     \\
MViTv2-L~(@312px)~\cite{b48}                                       & IN-21K    & 40     & 86.1          & 97.0          & 5×3   & 2828     \\ \midrule
\textit{\textbf{Web-scale image pretraining}}          &           &        &               &               &       &          \\
ViViT-H/16×2~\cite{b17}                                           & JFT-300M  & 32     & 84.8          & 95.8          & 4×3   & 8316     \\
TokenLearner-L/10~\cite{b27}                                      & JFT-300M  & -      & 85.4          & 96.3          & 4×3   & 4076     \\
CoVeR~\cite{b49}                                                 & JFT-3B  & -      & 87.2          & -             & 1×3   & -        \\ \midrule
\textit{\textbf{Web-scale language-image pretraining}} &           &        &               &               &       &          \\
ActionCLIP-B/16~\cite{b22}                                        & CLIP-400M & 32     & 83.8          & 96.2          & 10×3  & 563      \\
A6~\cite{b23}                                                     & CLIP-400M & 16     & 76.9          & 93.5          & -     & -        \\
EVL-ViT-B/16~\cite{b25}                                          & CLIP-400M & 16     & 83.6          & -             & 1×3   & 296      \\
EVL-ViT-L/14~\cite{b25}                                          & CLIP-400M & 16     & 87.0          & -             & 1×3   & 1350     \\
EVL-ViT-L/14@336px~\cite{b25}                                    & CLIP-400M & 32     & 87.7          & -             & 1×3   & 6068     \\
X-CLIP-B/16~\cite{b24}                                            & CLIP-400M & 16     & 84.7          & 96.8          & 4×3   & 287      \\
X-CLIP-L/14~(@336px)~\cite{b24}                                    & CLIP-400M & 16     & 87.7          & 97.4          & 4×3   & 3086     \\
AIM-ViT-L/14~\cite{yang2023aim}                                           & CLIP-400M & 16     & 87.3          & 97.6          & 1×3   & 1868     \\
ST-Adapter-ViT-L/14~\cite{pan2022st}                                    & CLIP-400M & 16     & 86.9          & 97.6          & 1×3   & 1375     \\
MTV-H~\cite{b34}                                                  & WTS       & 32     & {89.1} & {98.2} & 4×3   & 3705     \\ \midrule
ILA-ViT-B/32                                          & CLIP-400M & 8      & 81.3          & 95.0          & 4×3   & 40       \\
ILA-ViT-B/32                                    & CLIP-400M & 16     & 82.4          & 95.8          & 4×3   & 75       \\
ILA-ViT-B/16                                    & CLIP-400M & 8      & 84.0          & 96.6          & 4×3   & 149      \\
ILA-ViT-B/16                                    & CLIP-400M & 16     & 85.7          & 97.2          & 4×3   & 295      \\
ILA-ViT-L/14                                    & CLIP-400M & 8      & 88.0          & \textbf{98.1} & 4×3   & 673      \\
ILA-ViT-L/14@336px                              & CLIP-400M & 16     & \textbf{88.7} & 97.8          & 4×3   & 3130     \\ \bottomrule
\end{tabular}
}
\end{center}
\label{table:1}
\vspace{-0.2in}
\end{table*}

Our simple alignment implicitly introduces cross-frame cross-location interactions to the model, thus capturing semantically rich actions.
We reveal that the primitive pairwise interaction already contains sufficient information for modeling the complex temporal relations, which allows eliminating the costly temporal self-attention in video.
Therefore, there is no additional temporal modeling design in IST block.

\subsection{Training}
\label{sec:train}

The loss function of our framework consists of two parts.
The first part is the supervised prompt-enhanced similarity loss, where the cosine similarity between video representation $\mathbf{v}$ and text representation $\mathbf{c}$ is computed by:
\begin{equation}\small
\label{eq:losscos}
\begin{aligned}
     \mathbf{v}=&~\mathtt{Avg}(\mathtt{MSA}([\mathbf{z}^{(L)}_{1,cls},...,\mathbf{z}^{(L)}_{T,cls}])),\\
     &\mathtt{cos}(\mathbf{v},\mathbf{c})=\frac{<\mathbf{v},\mathbf{c}>}{||\mathbf{v}||\cdot||\mathbf{c}||}.
\end{aligned}
\end{equation}
Here $\mathtt{Avg}(\cdot)$ is the average pooling.
The model maximizes $\mathtt{cos}(\mathbf{v},\mathbf{c})$ if $\mathbf{v}$ and $\mathbf{c}$ are matched, otherwise minimizes it.

The second part is the alignment loss for aligning pairwise frames in each IST block.
Particularly, we align the mean-pooled feature, \emph{i.e.} the mutual information token $\mathbf{\hat{z}}^{(\ell)}_{t,mut}$ as in Eq. \eqref{eq:pool}, using the cosine similarity:
\begin{equation}\small 
    \label{eq:cosalign}
    \mathtt{cos}^{(\ell)}_{t}=\frac{<\mathbf{\hat{z}}^{(\ell)}_{t,mut},\mathbf{\hat{z}}^{(\ell)}_{t-1,mut}>}{||\mathbf{\hat{z}}^{(\ell)}_{t,mut}||\cdot||\mathbf{\hat{z}}^{(\ell)}_{t-1,mut}||},
\end{equation}
where $\mathtt{cos}^{(\ell)}_{t}$ is the similarity score for $t$-th frame pair in block $\ell$.
The loss function $l_{a}$ is formulated by summing up the similarity scores:
\begin{equation}\small 
    \label{eq:lossalign}
    l_{a}=-\sum^{L}_{\ell=1}\sum^{T}_{t=2}\mathtt{cos}^{(\ell)}_{t}.
\end{equation}
Finally, we optimize Eq. \eqref{eq:losscos} and Eq. \eqref{eq:lossalign} simultaneously with a loss weight parameter $\gamma$.

\begin{table*}[htbp]\small
\caption{Performance comparison with the state-of-the-arts on Something-Something-V2. The FLOPs per view of each method is reported.
}
 \vspace{-0.1in}
\begin{center}
\renewcommand\arraystretch{1.1}
\scalebox{1}{
\begin{tabular}{lcccccc}
\toprule
\textbf{Model}      & \textbf{Pretrain}             & \textbf{Frames} & \textbf{Top-1 Acc.} & \textbf{Top-5 Acc.} & \textbf{Views}          & \textbf{FLOPs~(G)} \\ \hline
ViViT-L~\cite{b17}             & IN-21K+K400                   & 16              & 65.4                    & 89.8                    & 1×3                     & 903               \\
TimeSformer-L~\cite{b13}       & IN-21K                        & 96              & 62.4                    & 81.0                    & 1×3                     & 2380              \\
TimeSformer-HR~\cite{b13}      & IN-21K                        & 16              & 62.2                    & 78.0                    & 1×3                     & 1703              \\
ATA~\cite{b28}                 & IN-21K                        & 32              & 67.1                    & 90.8                    & 4×3                     & 793               \\
MViTv1-B~\cite{b16}            & K400                          & 16              & 64.7                    & 89.2                    & 1×3                     & 70.5              \\
MViTv1-B~\cite{b16}            & K400                          & 32              & 67.1                    & 90.8                    & 1×3                     & 170               \\
Mformer-B~\cite{b14}           & IN-21K+K400                   & 16              & 66.5                    & 90.1                    & 1×3                     & 370               \\
Mformer-L~\cite{b14}           & IN-21K+K400                   & 32              & 68.1                    & 91.2                    & 1×3                     & 1185              \\
Mformer-HR~\cite{b14}          & IN-21K+K400                   & 64              & 67.1                    & 90.6                    & 1×3                     & 959               \\
X-CLIP-B/16~\cite{b24}                 & CLIP-400M                        & 8              & 57.8                    & 84.5                    & 4×3                     & 145               \\
AIM-ViT-B/16~\cite{yang2023aim}        & {CLIP-400M} & 8               & 66.4                    & 90.5                    & {1×3} & 208               \\
AIM-ViT-L/14~\cite{yang2023aim}        & {CLIP-400M} & 32              & 69.4                    & 92.3                    & {1×3} & 3836              \\
EVL-ViT-B/16~\cite{b25}        & CLIP-400M                     & 16              & 61.7                    & -                       & 1×3                     & 345               \\
EVL-ViT-L/14~\cite{b25}        & CLIP-400M                     & 32              & 66.7                    & -                       & 1×3                     & 3216              \\
EVL-ViT-L/14@336px~\cite{b25}  & CLIP-400M                     & 32              & 68.0                    & -                       & 1×3                     & 8090              \\
\midrule
ILA-ViT-B/16       & CLIP-400M                     & 8               & 65.0                    & 89.2                    & 4×3                     & 214               \\
ILA-ViT-B/16       & CLIP-400M                     & 16              & 66.8                    & 90.3                    & 4×3                     & 438               \\
ILA-ViT-L/14       & CLIP-400M                     & 8               & 67.8                    & 90.5                    & 4×3                     & 907               \\
ILA-ViT-L/14@336px & CLIP-400M                     & 16              &    70.2             &    91.8                 & 4×3                     & 3723              \\
\bottomrule
\end{tabular}
}
\end{center}
\label{table:2}
\vspace{-0.1cm}
\end{table*}

\section{Experiments}

We evaluate our method on two datasets: Kinetics-400~\cite{b46} and Something-Something-V2~\cite{b47}.
Four variants are considered, namely the ILA model based on ViT-B/32, ViT-B/16, ViT-L/14, and ViT-L/14@336, respectively.
We sparsely sample 8 or 16 frames to form a video clip, both in training and inference.
Additional implementation, hyperparameter details, and more experiments are provided in the supplementary materials.

\subsection{Main Results}

\noindent \textbf{Kinetics-400.} In Table \ref{table:1}, we report the performance of our proposed method on Kinetics-400. 
Comparisons with recent state-of-the-art approaches are listed, including methods with random initialization, pretrained on ImageNet-1k/21k pretraining, and pretrained with web-scale data.

Compared to methods pretrained on ImageNet \cite{deng2009imagenet}, ILA-ViT-L with 8 frames outperforms the best competitor MViTv2-L~\cite{b48} by $1.9\%$ in accuracy with 4$\times$ fewer FLOPs. 
We also observe ILA surpasses other baselines with large margins, \emph{e.g.}, Swin \cite{b15} and TimeSformer \cite{b13}.
It indicates the strong representations of the CLIP model, showing the great potential of large-scale visual-language pretraining.

In comparison with methods pretrained on web-scale images, \emph{e.g.} JFT-300M/3B, ILA exhibits significant advantages. Our ILA-ViT-L exceeds ViViT-H by $3.2\%$ with 12$\times$ less computation, and exceeds CoVeR by $0.8\%$. Note that CoVeR uses much more training data (3B images) compared to CLIP (400M image-text pairs).

In addition, when compared with the recent CLIP-based methods, ILA achieves the best performance.
ILA-ViT-B with 16 frames surpasses the typical CLIP-based model ActionCLIP-B by $1.9\%$ with 2$\times$ fewer FLOPs.
Moreover, our largest model outperforms the best competitors X-CLIP-L and EVL-L by $1\%$ with comparable or much less computation.
Though MTV-H performs a little higher ($89.1\%$) than ILA ($88.7\%$), it employs the WTS dataset that contains 70M video-text pairs with about 17B images, which are much larger than that in CLIP. 
The observations show that our alignment-based temporal modeling could capture more comprehensive motion clues than the insufficient temporal attention of X-CLIP and EVL, without increasing the computational burden.

\vspace{0.05in}
\noindent \textbf{Something-Something-V2.} Table \ref{table:2} reports the comparisons on SSv2.
This dataset focuses on the human object action recognition, in which the open domain semantics are limited.
We assume the rich textual representation of CLIP language branch can help less.
Therefore, we use the cross-entropy loss with one-hot labels, instead of the visual-text similarity loss in Eq. \eqref{eq:losscos}.
We also increase the number of convolution layers for better alignment.
Moreover, we freeze the weights of CLIP for stability.

SSv2 is a motion-heavy dataset and largely depends on temporal modeling.
Methods pretrained on CLIP usually produce weaker results compared to those pretrained on Kinetics-400~.
For example, X-CLIP-B only achieves $57.8\%$ in accuracy, while MViTv1-B produces much higher results ($64.7\%$) with similar computation.
Similarly, the result of EVL-ViT-B is also unsatisfactory ($61.7\%$).
This phenomenon can be attributed to three factors. 
(1) The temporal modeling in X-CLIP and EVL is insufficient. In pursuit of high efficiency, X-CLIP and EVL adopt frame-level or local temporal attention on top of the CLIP features, which inevitably harms the results.
(2) Tuning the weights of CLIP is very challenging, where small perturbations can easily prejudice the primal CLIP.
We assume the reason is that SSv2 is a dataset with relatively small semantics.
Even assigning a very small learning rate to CLIP weights and a large one to other weights, the model is still prone to encounter exploding gradients.
This phenomenon reduces the flexibility of parameter tuning, which leads to the insufficient training of the model.
(3) The pretraining on Kinetics can bring significant advantages compared to pretraining on CLIP data.

As shown in the table, ILA-ViT-B (8 frames) achieves a comparable $65.0\%$ with MViTv1-B, which is much higher than X-CLIP and EVL.
Moreover, ILA-ViT-L/14@336px obtains promising performance referring to $70.2\%$ on top-1 and $91.8\%$ on top-5. It outperforms EVL-ViT-L/14@336px by $2.2\%$ on top-1 with 2$\times$ fewer frames and over 2$\times$ fewer FLOPs.
It indicates that the proposed implicit alignment can comprehensively model the temporal information with a low computational cost.

\subsection{Ablation Study}

\noindent\textbf{Generalization to different backbones.} To demonstrate ILA is a versatile module and can be plugged into various backbones, we experiment with a CLIP-based model (EVL-ViT-B/16, 8frames~\cite{b25}) as well as an ImageNet-based architecture (TimeSformer-ViT-B/16, 8frames~\cite{b13}).  
For EVL, we insert our alignment into the CLIP backbone, while keep others unchanged.
For TimeSformer, we replace the temporal attention with the proposed alignment module.
The results are summarized in Table \ref{table:3}. 
The utilization of ILA results in a $0.6\%$ and $1.8\%$ performance gain for the CLIP-based and ImageNet-based backbones, respectively, demonstrating ILA is compatible with modern networks.

\begin{table}[h!]\small
\caption{Generalization ability of ILA on various visual backbones for Kinetics-400.
}
\vspace{-0.1in}
\begin{center}
\renewcommand\arraystretch{1.2}
\begin{tabular}{lccc}
\toprule
Model           & Pre-training & Acc. (\%)      & FLOPs         \\ 
\cmidrule(lr){1-1} \cmidrule(lr){2-2} \cmidrule(lr){3-4} \cmidrule(lr){4-4}
EVL~\cite{b25}             & CLIP-400M    & 82.9          & \textbf{150G} \\
EVL~+~ILA         & CLIP-400M    & \textbf{83.5} & 162G          \\ 
\cmidrule(lr){1-1} \cmidrule(lr){2-2} \cmidrule(lr){3-4} \cmidrule(lr){4-4}
TimeSformer~\cite{b13}     & IN-21K       & 78.0          & 196G          \\
TimeSformer~+~ILA & IN-21K       & \textbf{79.8} & \textbf{164G} \\ \bottomrule
\end{tabular}
\end{center}
\label{table:3}
\vspace{-0.1in}
\end{table}

\begin{table}[t!]\small
\caption{Effectiveness of implicit alignment on Kinetics-400. Average Pooling indicates forming the mutual information token in Eq. \eqref{eq:pool} without alignment.}
\vspace{-0.1in}
\begin{center}
\renewcommand\arraystretch{1.2}
\begin{tabular}{lcc}
\toprule
Model                   & Acc. (\%)      & FLOPs        \\ 
\cmidrule(lr){1-1} \cmidrule(lr){2-2} \cmidrule(lr){3-3}
Baseline                    & 79.8          & 37G          \\
X-CLIP~\cite{b24}                  & 80.4          & 39G          \\
CLIP + Divided ST Attention~\cite{b13} & 80.6          & 58G          \\
CLIP + Temporal Shift~\cite{b22}     & 80.1          & \textbf{37G} \\
CLIP + ATA~\cite{b28}                & 81.0          & 60G          \\
CLIP + Average Pooling          & 80.4          & 39G          \\ 
CLIP + ILA                & \textbf{81.3} & 40G          \\ \bottomrule
\end{tabular}
\end{center}
\label{table:4}
\vspace{-0.2in}
\end{table}

 \vspace{0.05in}
\noindent\textbf{Effectiveness of implicit alignment.} We compare ILA with ATA \cite{b28},  an alternative of patch alignment, and other temporal modeling approaches, \emph{i.e.} X-CLIP~\cite{b24}, Divided Spatio-Temporal Attention~\cite{b13}, Temporal Shift~\cite{b22}, and Average Pooling. The baseline is employing the loss in Eq. \eqref{eq:losscos} for CLIP without temporal modeling. Average Pooling indicates forming the mutual information token in Eq. \eqref{eq:pool} without alignment.
Table \ref{table:4} shows the comparison results.
We have the following observations: (1) ILA outperforms the baseline by $1.5\%$ in top-1 accuracy with minor additional computational cost. It indicates that ILA can promote CLIP for video tasks effectively. 
(2) Compared to ATA that uses patch-level movement for alignment with a cubic complexity,  ILA offers better results with nearly 2$\times$ fewer FLOPs through learning an implicit mask with a quadratic complexity. 
(3) ILA also outperforms other approaches like X-CLIP using temporal attention and temporal shifting, highlighting the effectiveness of ILA. 
(4) ILA achieves better results compared with average pooling, indicating that the improvement results from our implicit alignment instead of the pooling operation.

\begin{table}[h!]\small
\caption{Comparison of mutual information on Kinetics-400. MI~(EMD) refers to the average Wasserstein Distance between neighbouring frames.}
\vspace{-0.1in}
\begin{center}
\renewcommand\arraystretch{1.2}
\begin{tabular}{lcc}
\toprule
Model          & Acc. (\%) & MI~(EMD) \\ 
\cmidrule(lr){1-1} \cmidrule(lr){2-2} \cmidrule(lr){3-3}
Baseline           & 79.8   & 0.56    \\
X-CLIP~\cite{b24}           & 80.4    & 0.51    \\
CLIP + Divided ST Attention~\cite{b13} & 80.6   & 0.47    \\
CLIP + ATA~\cite{b28}       & 81.0   & 0.30    \\
CLIP + ILA       & \textbf{81.3}   & \textbf{0.13}    \\ \bottomrule
\end{tabular}
\end{center}
\label{table:6}
\vspace{-0.2in}
\end{table}

\vspace{0.05in}
\noindent\textbf{Comparison of mutual information.} In our work, we assume that ILA can enhance the mutual information between frames, thereby boosting the recognition performance. 
Here, we compare the mutual information of ILA in the last visual layer with other approaches.
In particular, we calculate the averaged Wasserstein Distance (\emph{i.e.} Earth Mover's Distance) \cite{arjovsky2017wgan} between adjacent frames which is negatively correlated to mutual information. 
Table \ref{table:6} presents the results.
We observe that models with additional alignment have lower Wasserstein Distance and higher performance, suggesting that the alignment can indeed correlate adjacent frames.

\vspace{0.05in}
\noindent\textbf{Impact of different aligning strategies.} ILA aligns two consecutive frames and here we experiment with the following alignment strategies: (1) Align-First: each frame is aligned with the first frame; (2) Align-Middle, each frame is aligned with the middle frame. 
We can observe in Table \ref{table:7} that anchor frame based alignments are inferior to adjacent alignment.
The reason may be that it is not reliable to align two frames that are too far away.

\begin{table}[htbp]\small
\caption{Ablation study of different aligning strategies on K-400. 
}
\vspace{-0.1in}
\begin{center}
\renewcommand\arraystretch{1.2}
\begin{tabular}{lcc}
\toprule
Aligning Strategy                       & Top1.~(\%)     & Top5.~(\%)     \\ 
\cmidrule(lr){1-1} \cmidrule(lr){2-2} \cmidrule(lr){3-3}
Align-First                   & 80.7          & 94.5          \\
Align-Middle                  & 80.8          & 94.6          \\
Adjacent frame & \textbf{81.3} & \textbf{95.0} \\ \bottomrule
\end{tabular}
\end{center}
\label{table:7}
\vspace{-0.2in}
\end{table}

\vspace{0.05in}
\noindent\textbf{Impact of inserting locations of alignment.} We divide the visual branch of ViT-B/32 (12 blocks) into 4 groups, each containing 3 blocks.
We plug our ILA into each group individually for exploring the impact of different inserting locations. Table \ref{table:8} shows the results. 
Per-block insertion of ILA outperforms the baseline CLIP by $0.6\%$, $0.7\%$, $0.6\%$ and $0.5\%$ in accuracy, respectively. 
We see that inserting ILA into shallow blocks performs slightly better than inserting it into deep ones, showing that aligning low-level features can encode more temporal clues.

\begin{table}[h!]\small
\caption{Comparisons of different inserting locations.}
\begin{center}
\renewcommand\arraystretch{1.2}
\begin{tabular}{lcc}
\toprule
Configuration & Acc.~(\%) & FLOPs \\ 
\cmidrule(lr){1-1} \cmidrule(lr){2-2} \cmidrule(lr){3-3}
None          & 79.8     & 37G   \\ 
\cmidrule(lr){1-1} \cmidrule(lr){2-2} \cmidrule(lr){3-3}
Block $1$-$3$     & 80.4     & 38G   \\
Block $4$-$6$     & 80.5     & 38G   \\
Block $7$-$9$     & 80.4     & 38G   \\
Block $10$-$12$    & 80.3     & 38G   \\ 
\cmidrule(lr){1-1} \cmidrule(lr){2-2} \cmidrule(lr){3-3}
ILA (Block $1$-$12$)           & 81.3     & 40G   \\ \bottomrule
\end{tabular}
\end{center}
\label{table:8}
\vspace{-0.1in}
\end{table}

\vspace{0.05in}
\noindent\textbf{Operators in alignment module.} 
To validate the effectiveness and efficiency of the 2D convolution module in ILA, we experiment with an alternative choice of window attention in Eq. \eqref{eq:conv}.
Table \ref{table:9} depicts the comparison results. 
It demonstrates that window attention requires high computational resources and is difficult to optimize, producing limited results.

\begin{table}[htbp]\small
\caption{Evaluation of different operators in Eq. \eqref{eq:conv}. We experiment with an alternative of window attention with size $3\times 3$, instead of the convolution.
}
\begin{center}
\renewcommand\arraystretch{1.2}
\begin{tabular}{lcc}
\toprule
Basic Operators          & Acc.~(\%) & FLOPs \\ 
\cmidrule(lr){1-1} \cmidrule(lr){2-2} \cmidrule(lr){3-3}
2D Convolution           & 81.3     & 40G   \\
Window Attention          & 80.8     & 114G  \\ 
\bottomrule
\end{tabular}
\end{center}
\label{table:9}
\vspace{-0.2in}
\end{table}

\vspace{0.05in}
\noindent\textbf{Impact of mutual information token.}
Here we discuss different approaches of exploiting the aligned features.
ILA employs a mutual information (MI) token by pooling and concatenation on aligned features.
Another choice is the element-wise addition between the frame and the aligned features.
In addition, one can also directly concatenate the tokens of aligned features to frame tokens, resulting 2$\times$ tokens in spatial attention.

The results are shown in Table \ref{table:10}. 
It can be observed that both element-wise addition and direct concatenation perform inferior to the ILA.
Furthermore, their inference latencies are much higher than ILA.
The plausible reason is that the aligned features are produced by simple mask weighting of frame features, thus containing much redundant information when performing addition or concatenation.
Meanwhile, the pooling operation can effectively remove such irrelevant information and boost the model performance.
\begin{table}[htbp]\small
\caption{Ablation study of mutual information (MI) token. ILA employs a MI token by pooling \& concatenation with aligned features.
Other choices include element-wise addition, or direct concatenation.}
\vspace{-0.2cm}
\begin{center}
\renewcommand\arraystretch{1.2}
\begin{tabular}{lccc}
\toprule
Implementation & Acc. & FLOPs & Latency~(ms) \\ 
\cmidrule(lr){1-1} \cmidrule(lr){2-2} \cmidrule(lr){3-3} \cmidrule(lr){4-4}
Element-wise Addition    & 80.2     & 44G   & 64.029                \\
Direct Concat. & 80.6     & 45G   & 58.548                \\
Pooling~\&~Concat. (ILA)      & 81.3     & 40G   & 47.075                \\ \bottomrule
\end{tabular}
\end{center}
\label{table:10}
\vspace{-0.2in}
\end{table}

\begin{figure*}[t!]
\begin{center}
\includegraphics[width=0.8\linewidth]{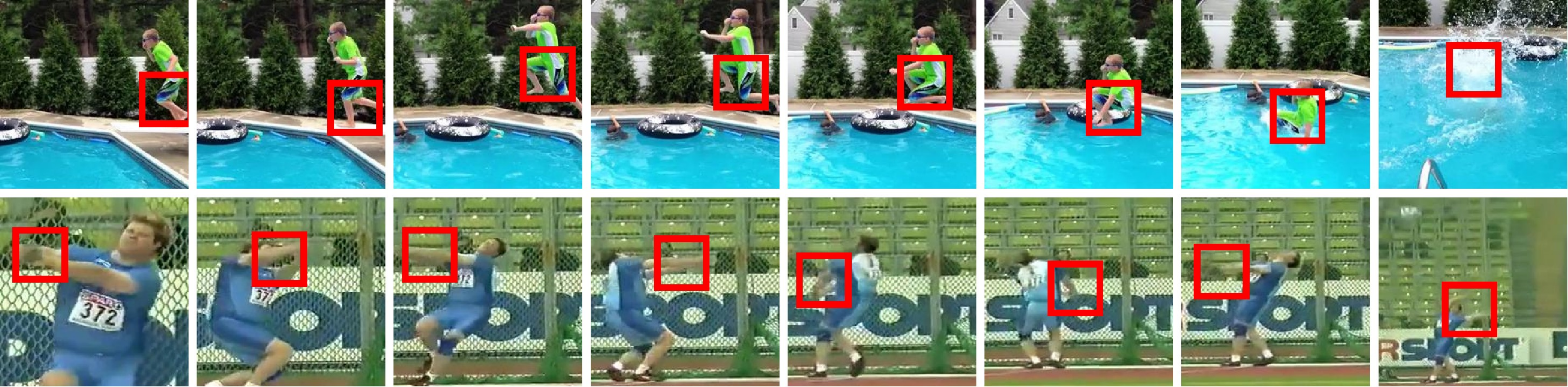}
\end{center}
  \caption{Visualization of mutual information over time. We draw tracking borders around each interactive point in a fixed size, which the tracking borders along temporal dimension depict the temporal corresponding mutual information captured by ILA.}
\label{fig:track}
\end{figure*}

\noindent\textbf{Impact of CLIP initialization.} In order to eliminate the influence of CLIP weights, we initialize the weights of ILA with ViT-B/16 pretrained on IN-21K, as well as removing the text branch and using the one-hot labels instead, which is same to the Swin. The results on K400 are shown on Table \ref{table:11}. ILA outperforms Swin by $0.6\%$ on Top-1 accuracy under the same pretraining setting, indicating the superiority of ILA. The results demonstrate that our proposed model obtains the promising performance due to ILA itself instead of language encoder of CLIP or CLIP pretraining weights. 

\begin{table}[htbp]\small
  \caption{Comparison results between ILA and Swin under the same weight parameters initialization.}
  \vspace{-0.2cm}
  \begin{center}
  \renewcommand\arraystretch{1.2}
  \begin{tabular}{lcccc}
    \toprule
    Model           & Pretraining~~ & R-1~ & R-5~ & Views \\ \midrule
Swin-B/32f~~ & IN-21K   & 82.7      & 95.5      & 4×3   \\
ILA (ViT-B/16-32f)  & IN-21K   & \textbf{83.3}      & \textbf{95.8}      & 4×3   \\ \bottomrule  
  \end{tabular}
  \end{center}
  \label{table:11}
  \vspace{-0.2in}
\end{table}

\noindent\textbf{Ablation of text representation tuning.} In terms of ILA architecture, We follow X-CLIP to adopt the video-conditioned text representation tuning. For fair comparisons, we run an additional ablation by removing it on K400 with ViT-B/32-8f. In Table \ref{table:12}, we can see that ILA still performs better than X-CLIP even without the video-specific tuning.

\begin{table}[htbp]\small
  \caption{Ablation results of text representation tuning.}
  \begin{center}
  \renewcommand\arraystretch{1.2}
  \begin{tabular}{lcc}
    \toprule
    Model  & w/o~video-specific text & w/~video-specific text \\ \midrule
X-CLIP & 79.6                         & 80.4                        \\
ILA    & \textbf{80.8}                         & \textbf{81.3}                        \\ \bottomrule
  \end{tabular}
  \end{center}
  \label{table:12}
  \vspace{-0.2in}
\end{table}

\subsection{Additional Comparison Results}

\noindent\textbf{Zero-shot results on SSv2.} We train ILA and other CLIP-based competitors on K400 with ViT-B/16-8f and evaluate on SSv2 in a zero-shot setting, by training an additional linear classification layer. The results are depicted on Table \ref{table:13}. We see that ILA outperforms two typical competitors X-CLIP and EVL by $5.8\%$ and $8.7\%$ on Top-1 respectively, highlighting the effectiveness of ILA.

\begin{table}[htbp]\small
  \caption{Zero-shot results on SSv2.}
  \begin{center}
  \renewcommand\arraystretch{1.2}
  \begin{tabular}{llcc}
    \toprule
    Model  & Pretraining   & Top1 (\%)      & Top5 (\%)      \\ \midrule
    X-CLIP~\cite{b24} & CLIP-400M & 38.1          & 68.1          \\
    EVL~\cite{b25}    & CLIP-400M    & 35.2          & 65.4          \\
    AIM~\cite{yang2023aim}    & CLIP-400M    & 39.1          & 68.7          \\ 
    ILA    & CLIP-400M    & \textbf{43.9} & \textbf{71.8} \\
    \bottomrule
  \end{tabular}
  \end{center}
  \label{table:13}
  \vspace{-0.2in}
\end{table}

\noindent\textbf{Performance on additional benchmark.} We evaluate ILA, X-CLIP~\cite{b24} and EVL~\cite{b25} on the temporal understanding benchmark~\cite{sevilla2021only} which is without static biases. The results are shown in Table \ref{table:14}. All models are pretrained on CLIP-400M (ViT-B/16-8f). Top-1 (T) and Top-1 (S) refer to the traditional accuracy on Temporal-50 and Static-50, respectively. TS refers to the relative gain of the model on temporal classes compared to static ones (Temporal score). In Table \ref{table:14}, we can see that our ILA outperforms X-CLIP by $6.0\%$ on Top-1 and by $3.8\%$ on temporal score. ILA also exceeds the best temporal score on RGB modality (5.2\%, R3D) in the paper~\cite{sevilla2021only}. The result highlights the effectiveness of ILA on the temporal understanding benchmark.  

\begin{table}[htbp]\small
  \caption{Performance comparisons on additional benchmark without static biases.}
  \begin{center}
  \renewcommand\arraystretch{1.2}
  \begin{tabular}{lcccc}
    \toprule
Model  & Top-1~(T)      & Top-5~(T)      & Top-1~(S)      & TS \\ \midrule
X-CLIP & 75.9          & 94.1          & 73.6          & 2.3            \\
EVL    & 70.7          & 92.5          & 68.3          & 2.4            \\
ILA    & \textbf{81.9} & \textbf{97.4} & \textbf{75.8} & \textbf{6.1}           \\ \bottomrule 
  \end{tabular}
  \end{center}
  \label{table:14}
  \vspace{-0.2in}
\end{table}

\subsection{Visualization of mutual information}
\noindent We visualize the interactive point at each video frame by drawing the bounding boxes centered at the points, as illustrated in the Figure \ref{fig:track}. The bounding boxes indicate the regions with rich favorable mutual information, where the moving boxes show the potential of abilities on tracking moving objects.

\section{Conclusion}

We introduced Implicit Learnable Alignment (ILA), a novel temporal modeling method for video recognition. ILA performs frame alignment so as to encode motion information in lieu of the widely used temporal attention operation. Particularly, ILA employs only an implicit and coarse feature alignment via a weighting mask. 
By finding the active interaction position in the frame, the mask is generated with higher weights around that position, and lower weights on others. Extensive experiments demonstrates the effectiveness and efficiency of ILA, showcasing its promising adaption ability to CLIP and compatibility with modern visual backbones.

\noindent\textbf{Acknowledgement} This project was supported by NSFC under Grant No. 62032006  and No. 62102092.

{\small
\bibliographystyle{ieee_fullname}

}

\appendix

\section*{Appendix}

\section{Implementation Details of ILA}

\noindent\textbf{Training Details.} The experiments are conducted on 8 NVIDIA 32G V100 GPUs. The training configuration is listed in Table \ref{table:imple}. It is worth noting that our sampling strategies for Kinetics-400 and Something-Something-V2 are different during the training phase. We implement the sparse sampling strategy on Kinetics-400. For SSv2, we uniformly sample the entire video at predefined temporal intervals without group division. 
In term of the training on Kinetics-400, the base learning rate indicates the learning rate of the original CLIP parameters. The learning rate for other additional parameters is 10$\times$ larger than the base learning rate. 
In term of the training on SSv2, we exclude the prompt branch and freeze the weights of CLIP visual branch for training stability.
Thus the base learning rate is used for the rest parameters.

\begin{table}[htbp]\small
\caption{Default implementation details of our method.}
 \vspace{-0.0in}
\begin{center}
\renewcommand\arraystretch{1.1}
\scalebox{0.87}{
\begin{tabular}{lcc}
\toprule
Training Configuration                 & Kinetics-400    & Something-Something v2   \\ \midrule
\textit{\textbf{Optimisation}}         &                 &                          \\
Optimizer                              & \multicolumn{2}{c}{AdamW}                  \\
Optimizer betas                        & \multicolumn{2}{c}{(0.9,0.98)}             \\
Batch size                             & \multicolumn{2}{c}{256}                    \\
Learning rate schedule                 & \multicolumn{2}{c}{Cosine}                 \\
Learning warmup epochs                 & \multicolumn{2}{c}{5}                      \\
Base learning rate                     & 8e-6            & 5e-4                     \\
Minimal learning rate                  & 8e-8            & 5e-6                     \\
training steps                         & 50000           & 30000                    \\ \midrule
\textit{\textbf{Data augmentation}}    &                 &                          \\
RandomFlip                             & \multicolumn{2}{c}{0.5}                    \\
MultiScaleCrop                         & \multicolumn{2}{c}{(1, 0.875, 0.75, 0.66)} \\
ColorJitter                            & \multicolumn{2}{c}{0.8}                    \\
GrayScale                              & \multicolumn{2}{c}{0.2}                    \\
Label smoothing                        & \multicolumn{2}{c}{0.1}                    \\
Mixup                                  & \multicolumn{2}{c}{0.8}                    \\
Cutmix                                 & \multicolumn{2}{c}{1.0}                    \\ \midrule
\textit{\textbf{Other regularisation}} &                 &                          \\
Weight decay                           & 0.003           & 0.01                     \\ \bottomrule
\end{tabular}
}
\end{center}
\label{table:imple}
\vspace{-0.1in}
\end{table}

\noindent \textbf{Convolution Module in SSv2.} 
In SSv2, we increase the number of convolution layers in alignment.
Particularly, two additional 3$\times$3 convolution layers plus batch normalization and ReLU are added.
In comparison to the original convolution module, it can bring 0.6\% improvement on top-1 accuracy.

\section{Complexity of ILA} 
We analyze various temporal modeling methods (Spatial Attention~\cite{b13}, Joint Attention~\cite{b13}, Divided ST Attention~\cite{b13}, ATA~\cite{b28}, X-CLIP~\cite{b24} and our proposed ILA) in terms of complexity, as shown in Table \ref{table:complex}. 
The complexity of our alignment process is $O(Thwk^{2}d)$ due to the 2D convolution-based operations. 
The complexity of the whole ILA consists of the implicit alignment $O(Thwk^{2}d)$ and the spatial attention $O(Th^{2}w^{2}d)$. 
In terms of Joint Attention and Divided Spatiotemporal Attention, Joint Attention requires more computational memory since it takes all patches into consideration. 
Divided ST Attention only considers the temporal attention along the time axis.
In terms of ATA, ATA is based on Hungarian Algorithm whose complexity is $O(N^{3})$. In practice, the complexity of Hungarian matching is $O(Th^{3}w^{3}d)$ in video domain. 
Moreover, ATA requires additional temporal attention with complexity $O(T^{2}hwd)$. 
X-CLIP adopts a frame-level temporal attention with complexity $O(T^{2}d)$, which however obtains suboptimal result.
We can observe that our proposed ILA can have better performance in low complexity.

\section{Qualitative Analysis}

\begin{table*}[t!]
\caption{Complexities of different methods, with results on Kinetics-400. $T$, $h$, $w$, $d$, and $k$ refer to temporal size, spatial height of input, spatial width of input, channel depth of input, and kernel size of convolution, respectively.}
\begin{center}
\begin{tabular}{lccc}
\hline
Temporal Modeling    & Complexity                                                                                                       & Acc.(\%) & FLOPs \\ \hline
Spatial Attention~\cite{b13}    & $O(Th^{2}w^{2}d)$                                                                    & 79.8     & 37G   \\
Joint Attention~\cite{b13}      & $O(T^{2}h^{2}w^{2}d)$                                                & 80.4     & 71G       \\
Divided ST Attention~\cite{b13} & $O(T^{2}hwd+Th^{2}w^{2}d)$                                            & 80.6     & 58G   \\
ATA~\cite{b28}                  & $O(Th^{3}w^{3}d+T^{2}hwd+Th^{2}w^{2}d)$ & 81.0     & 60G   \\
X-CLIP~\cite{b24}                  & $O(T^{2}d+Th^{2}w^{2}d)$ & 80.4     & 39G   \\
ILA                  & $O(Thwk^2d+Th^{2}w^{2}d)$                         & 81.3     & 40G   \\ \hline
\end{tabular}
\end{center}
\label{table:complex}
\end{table*}

\begin{figure*}[htbp]
\begin{center}
\includegraphics[width=0.8\linewidth]{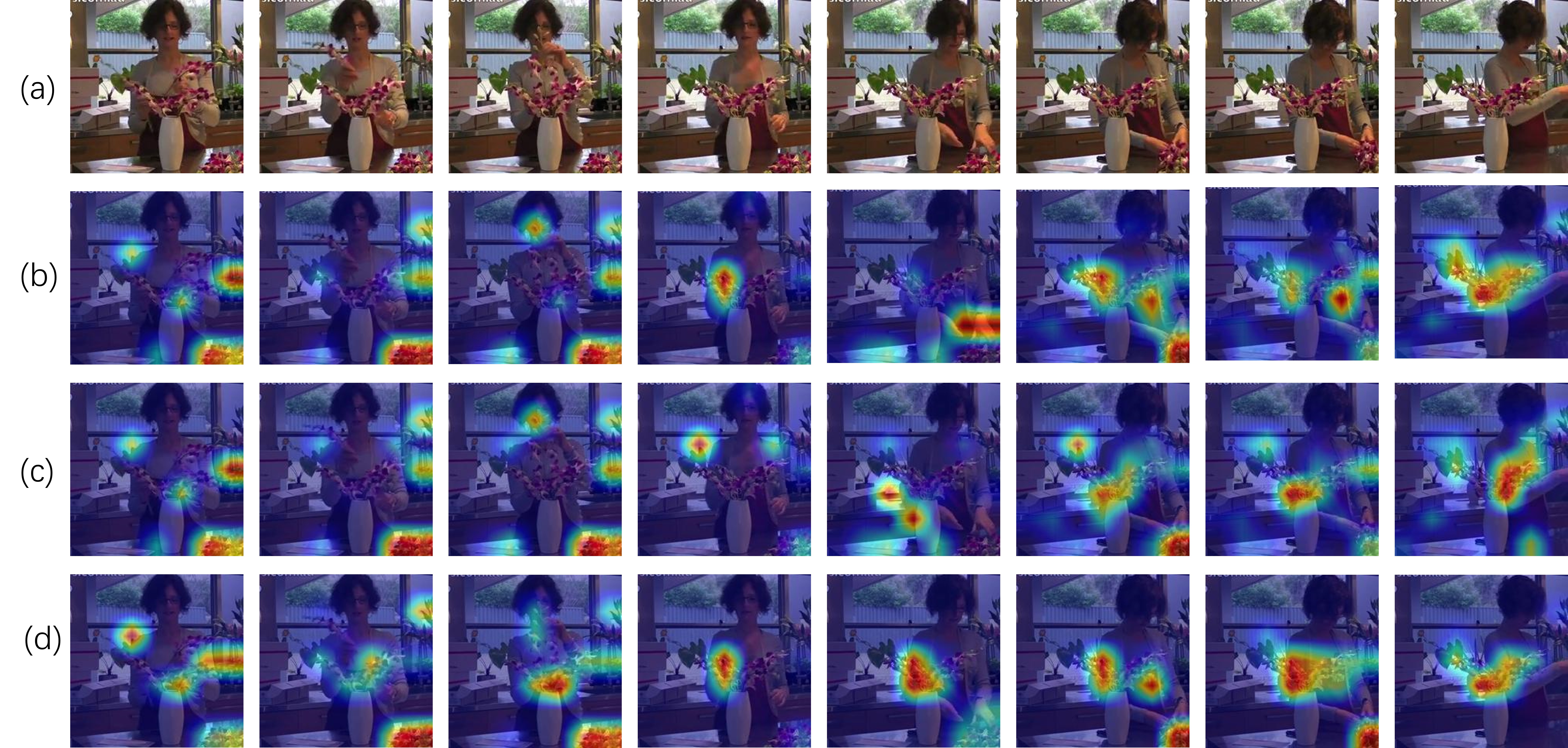}
\end{center}
  \caption{Visualization of intermediate feature map of different temporal modeling approaches on Kinetics-400. (a) refers to raw frames. (b), (c) and (d) refer to Divided ST Attention, ATA and ILA respectively.}
\label{fig:heatmap_middle}
\end{figure*}

\begin{figure*}[htbp]
\begin{center}
\includegraphics[width=0.8\linewidth]{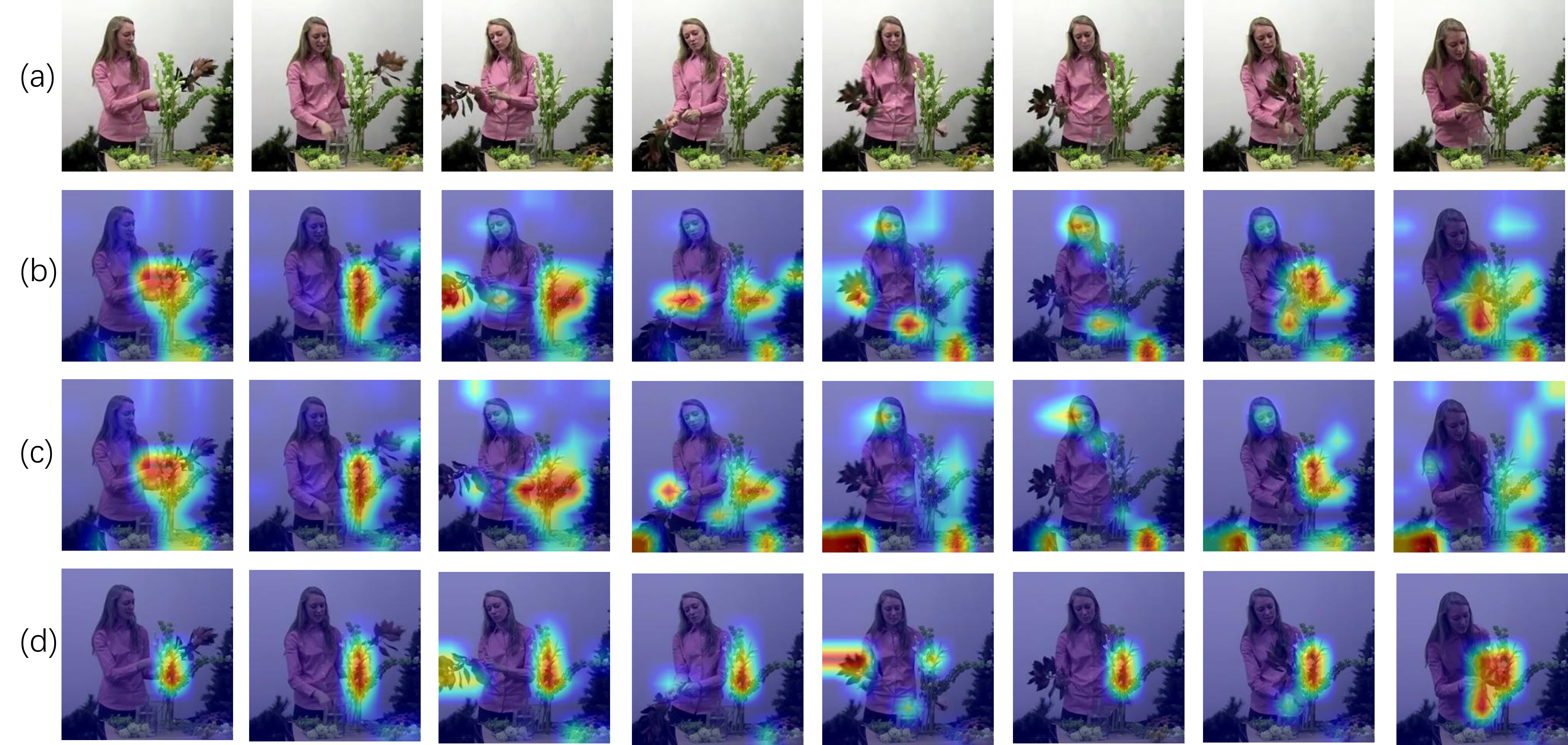}
\end{center}
  \caption{Visualization of the last feature map of different temporal modeling approaches on Kinetics-400. (a) refers to raw frames. (b), (c) and (d) refer to Divided ST Attention, ATA and ILA respectively.}
\label{fig:heatmap_final}
\end{figure*}

In order to investigate the quality of three temporal modeling approaches (Divided ST Attention~\cite{b13}, ATA~\cite{b28}, and ILA), we visualize their intermediate and last feature maps respectively, as shown in Figure \ref{fig:heatmap_middle} and Figure \ref{fig:heatmap_final}.
According to the illustrations, all three approaches capture the static semantic features, such as static flowers on the desk. 
Moreover, our proposed ILA pays more attention to the action area of arranging flowers (\emph{e.g.} the 5-th frame in the last row of Figure \ref{fig:heatmap_final}) instead of the static flowers on the desk. 
It indicates that our ILA can leverage the learnable mask to achieve implicit temporal modeling, focusing on the vital motion region.
For divided ST attention, the model prefers to focus on static object instead of significant actions. 
While in ATA, the model attempts to concentrate on discontinuous regions with inaccurate positions.
The plausible reason is that ATA utilizes patch movement-based alignment, which may destroys the continuity of semantic distribution. 

\section{Key differences between ILA and ATA} ATA adopts an explicit patch-level alignment with Hungarian matching, aiming at modeling temporal attention within aligned patches, which has poor efficiency due to the frame-by-frame serial alignment. Our ILA is fundamentally different as we utilize learnable masks to obtain implicit and coarse semantic-level alignment, which attempts to enhance favorable mutual information and can be performed in parallel with high efficiency. 

It exits three fundamental different aspects. First, ATA can only align the collection of frames representations in serial frame-by-frame mode due to the limitation of KMA, while our ILA can utilize learnable masks to align semantical correspondences between two neighboring frames in parallel resulting in faster inference. Second, the complexity of ATA is $O(N^{3})$ and ATA is unlearnable resulting in difficult optimization. Computational complexity of ILA is $O(N^{2})$. Third, the core idea of ATA is to implement KMA algorithm to find out the optimal patch-level movement scheme capturing temporal correspondences, while the core idea of our ILA is to utilize specific masks to suppress irrelevant redundant information and enhance task-related mutual information among frames resulting in implicit alignment. Therefore, ATA still preserve the original irrelevant redundant information, while our ILA has the suppression of irrelevant redundant information due to principle of masks.

\end{document}